\documentclass{bmvc2k}

\title{Cascaded Cross MLP-Mixer GANs for Cross-View Image Translation}

\addauthor{Bin Ren}{bin.ren@unitn.it}{1}
\addauthor{Hao Tang}{hao.tang@vision.ee.ethz.ch}{2}
\addauthor{Nicu Sebe}{niculae.sebe@unitn.it}{1}

\addinstitution{
 DISI\\
 University of Trento
}
\addinstitution{
 Computer Vision Lab\\
 ETH Zurich
}

\runninghead{Student, Prof, Collaborator}{BMVC Author Guidelines}

\usepackage{mathrsfs}
\usepackage{bbm}
\usepackage{amsmath,amssymb,textcomp,subfigure,multirow,upgreek}
\usepackage{booktabs}
\usepackage{color,mdwlist}
\usepackage{graphicx,fontawesome}
\usepackage{enumitem}

\usepackage{enumitem}
\setitemize{noitemsep,topsep=0pt,parsep=0pt,partopsep=0pt}

\usepackage{caption}
\begin{document}

\maketitle

\begin{abstract}
It is hard to generate an image at target view well for previous cross-view image translation methods that directly adopt a simple encoder-decoder or U-Net structure, especially for drastically different views and severe deformation cases. To ease this problem, we propose a novel two-stage framework with a new Cascaded Cross MLP-Mixer (CrossMLP) sub-network in the first stage and one refined pixel-level loss in the second stage. In the first stage, the CrossMLP sub-network learns the latent transformation cues between image code and semantic map code via our novel CrossMLP blocks. Then the coarse results are generated progressively under the guidance of those cues. Moreover, in the second stage, we design a refined pixel-level loss that eases the noisy semantic label problem with more reasonable regularization in a more compact fashion for better optimization. Extensive experimental results on Dayton~\cite{vo2016localizing} and CVUSA~\cite{workman2015wide} datasets show that our method can generate significantly better results than state-of-the-art methods. The source code and trained models are available at \url{https://github.com/Amazingren/CrossMLP}.
\end{abstract}
\section{Introduction}
\label{sec:intro}
Cross-view image translation is a task that aims at synthesizing new images at the target view from the source view. It has achieved lots of interest due to it can be applied to city scene synthesis~\cite{zhou2016view} and 3D object translation~\cite{yang2015weakly} from both the computer vision and virtual reality communities~\cite{tatarchenko2016multi,kim2017learning,zhu2018generative,regmi2018cross,zhai2017predicting,huang2017beyond,park2017transformation,zhou2016view,yang2015weakly,tang2019multi}. This task is usually investigated by Convolutional Neural Networks (CNNs) based encoder-decoder structure~\cite{zhou2016view,yang2015weakly} or Generative Adversarial Networks (GANs) based methods~\cite{park2017transformation}. However, these methods were all designed to generate new view images that share a degree of overlap on both appearances and geometry information with the original ones.
Unlike previous methods mentioned above, Regmi \textit{et al.}~\cite{regmi2018cross} proposed a conditional GAN which jointly learns the generation in both the image domain and its corresponding semantic domain, and the semantic predictions are further utilized to supervise the image generation. To further improve the performance, Tang \textit{et al.}~\cite{tang2019multi} proposed SelectionGAN, a coarse-to-fine method which built on three complicated U-Net generators and a selection module. 

\begin{figure}[!t] \small
	\centering
	\includegraphics[width=0.9\linewidth]{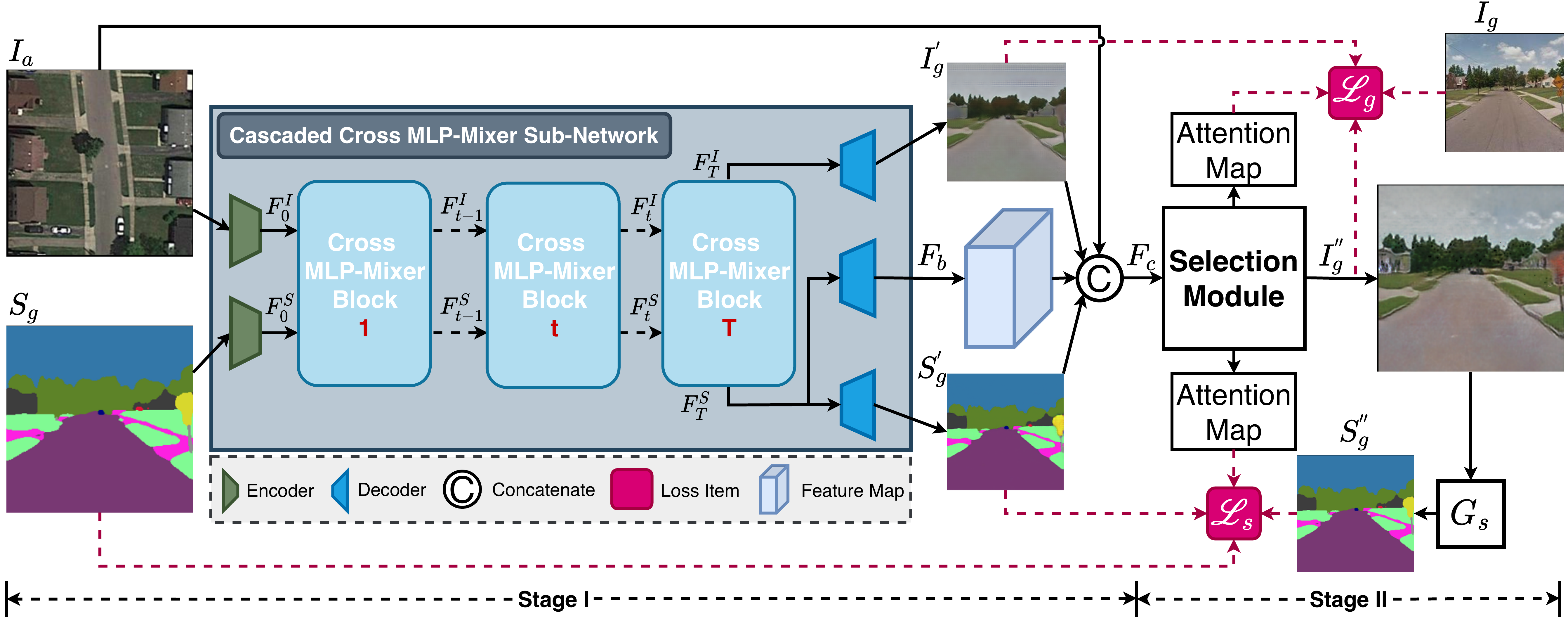}
	\caption{Overview of the proposed Cascaded Cross MLP-Mixer GANs.}
	\label{fig:framework}
	\vspace{-0.4cm}
\end{figure}

Although these two methods performed interesting explorations, we observe that there are three main factors that can account for the reason why the quality of their generated images is still far from satisfactory. Firstly, previous methods generate the target view image just in an one-shot manner directly, which ignores some important details during the translation. Secondly, the interaction between the source view image and the target view semantic map lacks a deep exploration since it is critical for a model to learn a precise mapping pattern, especially for the case that there is a drastic gap between source view and target view in both appearance and geometry. Thirdly, the inevitable noisy semantic label problem needs to be further addressed since it will mislead the image generation because the semantic label maps are not accurate enough for all the pixels when they are usually produced from pretrained semantic models.  

To this end, we propose a two-stage framework with a novel Cascaded Cross MLP-Mixer (CrossMLP for short) sub-network in the first stage (see Figure~\ref{fig:framework}) and a refined pixel-level loss in the seond stage. Within the CrossMLP sub-network, we stack several novel Cross MLP-Mixer (CrossMLP) blocks to produce a coarse image and a coarse semantic map at target view. This procedure is just like a human drawer is painting a picture~\cite{gregor2015draw}, progressively. What's more, to dig out the latent translation cues between the source view image and the target view semantic map, we design the CrossMLP module (see Figure~\ref{fig:crossblock}) within each CrossMLP block in the generator of a GAN. It interactively model the mutual latent long-range dependencies which consist of both the appearance and the structure information within each view, and the latent correspondences between two inputs can also be captured via its two consecutive multi-layer perceptrons (MLP). Then the latent cues are used to guide the transformation from source view to target view.

In the second stage, we propose the refined pixel-level loss. On one hand, it can ease the noisy semantic label problem by more reasonable regularization to the overall optimization loss. On the other hand, the formulation of our refined pixel-level loss becomes more compact with lower redundancy. Consequently, the generated images at target views become more realistic. Finally, extensive experimental results demonstrate that our method achieves new state-of-the-art results on both Dayton~\cite{vo2016localizing} and CVUSA~\cite{workman2015wide} datasets.

The contributions of this paper are summarized as follows,
\begin{itemize}
	\item We propose a two-stage framework that can progressively generate the target view image via the proposed CrossMLP blocks. To the best of our knowledge, the proposed model is the first MLP-based cross-view image translation framework.
	\item We propose a novel CrossMLP module to dig out the latent mapping cues between the source view image and the target view semantic map. The latent cues can guide the cross-view translation with better geometry structure and appearance information.
	\item We propose a refined pixel-level loss, which can reduce the redundancy of loss items and ease the noisy semantic label problem in the cross-view translation task.
	\item Extensive experiments demonstrate the effectiveness of our method and show new state-of-the-art results on most of the evaluation metrics for both datasets.
\end{itemize}

\section{Related Work}
\label{sec:relatd}
\noindent \textbf{Generative Adversarial Networks (GANs).} As one of the most used techniques of the generative model, GANs are a unique approach for learning desired data distribution to generate new samples~\cite{goodfellow2014generative}. Generally, there is a generator with a discriminator that exists in GANs. The former tries to produce photo-realistic target images to fool the latter, while the latter tries its best to figure out whether a sample is real or fake. Though lots of works have shown their capability in generating realistic-looking images~\cite{brock2018large,shaham2019singan,karras2019style}, it is still challenging for vanilla GANs to produce images in a controlled manner. As an extension solution to this problem, conditional GANs (CGANs) are proposed by incorporating conditional information~\cite{goodfellow2014generative} such as class labels~\cite{esser2018variational,wang2004image,zakharov2019few}, attention maps~\cite{kim2019u, tang2021attentiongan}, human skeleton~\cite{tang2020bipartite,zhu2019progressive,tang2020xinggan}, and semantic maps~\cite{park2019semantic,tang2020local,tang2021layout,liu2021cross,liu2020exocentric}. In this paper, our work focuses on cross-view image generation with semantic maps as conditional information.

\noindent \textbf{Image-to-Image Translation} aims to learn a parametric mapping between input and output data. Isola \textit{et al.} realized this by its supervised Pix2pix method to learn a translation function between input and results based on CGANs. Zhu \textit{et al.} \cite{zhu2017unpaired} introduced CycleGAN, which using the cycle-consistency loss to solve the unpaired image translation problem. In addition, there are lots of works like \cite{chen2018attention,xu2018attngan,ma2018gan,mejjati2018unsupervised} that all adopted attention mechanism to improve the generation performance. Moreover, based on the self-attention strategy, Transformer based methods have become popular relied on their ability for modeling global long-range dependencies in vision tasks compared to conventional CNN methods recently~\cite{jiang2021transgan,ren2021cloth}. However, due to Transformer is built on the self-attention mechanism, when the size of the feature map increases, the computing and memory overheads increase quadratically. 
To ease this problem, inspired by MLP-Mixer \cite{tolstikhin2021mlp} which closely resembles the Transformer model in an efficient way and has shown its powerful ability on discriminative vision tasks~\cite{dosovitskiy2020image}. We propose the novel CrossMLP in this paper which is based entirely on multi-layer perceptrons (MLPs) for both spatial locations and feature channels module to address the generative task. Therefore it can model the long-range dependencies via a more simple alternative in GAN.

\noindent \textbf{Cross-View Transformation.}
Most existing works on cross-view transformation mainly focuse on synthesizing novel views of the same object, such as chairs, tables, or cars~\cite{dosovitskiy2017learning,tatarchenko2016multi,choy20163d}. Another group of works \cite{yin2018novel,zhou2016view} explore the cross-view scene image translation problem in which there exists a large degree of overlapping in both appearances and geometry structures. However, when facing the drastically different view and severe formation cases which involve huge geometry mismatch, large-scale uncertainty, and obvious appearance difference, it's really challenging for existing methods. Zhai \textit{et al.}~\cite{zhai2017predicting} firstly proposed a CNN-based network for generating panoramic ground-level images from aerial images at the location. Krishna \textit{et al.}~\cite{regmi2018cross} proposed a GAN-based structure X-Fork and an X-Seq to address the aerial to street-view image translation via an extra semantic segmentation map. To improve the performance, Tang \textit{et al.}~\cite{tang2019multi} proposed SelectionGAN, a coarse-to-fine framework based on three U-Net generators and a Selection module. However, nearly all the methods mentioned above generate the results in a hard one-shot way, the final generated results are still far from satisfactory due to the drastic difference between source and target views. 
Hence, we propose a novel progressive generation fashion via several cascaded CrossMLP blocks for this task. This strategy allows our method can interactively produce complex images at target view with much better generation quality step by step.

\section{Cascaded Cross MLP-Mixer GANs}
\label{sec:method}

\subsection{Overview}

\noindent \textbf{Notations.} 
The overall structure of the proposed Cascaded Cross MLP-Mixer GAN is depicted in Figure~\ref{fig:framework}. There are two stages in this framework. The first one takes as input both one image $I_a$ at source view and one semantic map $S_g$ at the target view, aiming to progressively produce a coarse image
$I_{g}^{'}$ at target view, a coarse target semantic map $S_{g}^{'}$, and a feature map $F_{b}$ which contains the latent transformation cues between $I_{a}$ and $S_{g}$. The second stage firstly takes $F_c$, a combination feature originated from $I_a$, $I_{g}^{'}$, $S_{g}^{'}$, and $F_{b}$, as input. Then $F_c$ is sent to the selection module proposed by Tang \textit{et al.}~\cite{tang2019multi}, note that this module is just used for producing the final cross-view image $I_{g}^{''}$ and uncertainty maps $U$ to multiple optimization losses, which is our secondary focus. Moreover, $I_{g}^{''}$ is pass through $G_s$, a U-net based structure, for recovering a more detailed semantic map $S_g^{''}$. Finally, all image related items are used to construct the refined pixel-level image loss $\mathscr{L}_{g}$ while all semantic maps related items are used to construct the refined pixel-level semantic map loss $\mathscr{L}_{s}$, through which a more fine-grained synthesized cross-view image can be produced.

\noindent \textbf{Encoder and Decoder.} As shown in Figure~\ref{fig:framework}, the input image $I_a$ from the source view and the semantic map $S_g$ from one target view are firstly encoded by two similar feature extraction encoders that consist of $N$ down-sampling convolutional layers ($N$=2 in our case). On the output side of our proposed cascaded MLP-Mixer generation, both the coarse output image $I_g^{'}$ and the coarse output semantic map $S_g^{'}$ at the target view are recovered by two similar decoders from feature $F_{T}^{I}$ and $F_{T}^{S}$ via $N$ deconvolutional layers. What's more, for the intermediate feature map $F_b$, we recover it from $F_{T}^{S}$ only via $N/2$ deconvolutional layers.

\subsection{Cascaded Cross MLP-Mixer Sub-Network}
\label{sec:MLP-Mixer}
\begin{figure}[!t] \small
	\centering
	\includegraphics[width=0.9\linewidth]{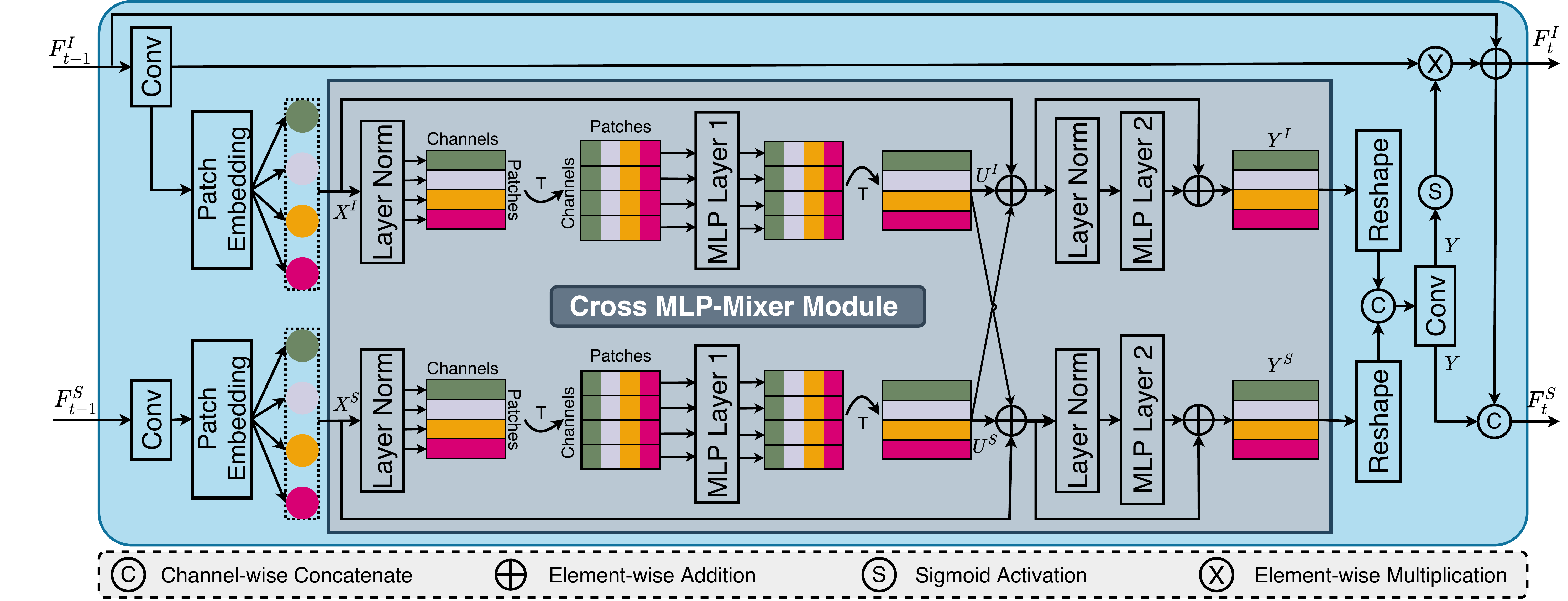}
	\caption{Architecture of the proposed Cross MLP-Mixer block.}
	\label{fig:crossblock}
		\vspace{-0.4cm}
\end{figure}

As the core of the generator in the first stage, the cascaded cross MLP-Mixer sub-network is consisting of several CrossMLP blocks shown in Figure~\ref{fig:framework}. Starting from the initial image code $F_{0}^{I}$ and semantic map code $F_{0}^{S}$. The CrossMLP sub-network progressively updates these two codes through a sequence of CrossMLP blocks. Then the final image code $F_{T}^{I}$ is taken to decode the coarse output image $I_{g}^{'}$ at target view. While the final semantic code $F_{T}^{S}$ is token to decode for both the feature map $F_{b}$ and the semantic map $S_{g}^{'}$. All CrossMLP blocks have identical structures, and each carries out one step of update. Consider the $t$-th block depicted in Figure~\ref{fig:crossblock}, whose inputs
are $F_{t-1}^{I}$ and $F_{t-1}^{S}$. Each block comprises two pathways, called the image pathway and the semantic map pathway. With deep interactions via CrossMLP module, these two pathways update $F_{t-1}^{I}$ and $F_{t-1}^{S}$ to $F_{t}^{I}$ and $F_{t}^{S}$, respectively. In the following, we describe the detailed update process in three parts and justify their designs.

\noindent \textbf{CrossMLP Module.} From a basic level, the cross-view image translation is to moving patches from the location induced by the source view image to the location induced by the target view semantic map. So how to let the translation knows the latent cues that where to sample source patches and where to put target patches is extremely significant. Therefore, we design the novel CrossMLP module for modeling the global long-range relation between source view image code and target view semantic map code to realize such cues.

Modern deep vision architectures generally consist of layers that mix features (i) at a given spatial location, (ii) between different spatial locations, or both at once~\cite{tolstikhin2021mlp}. CNNs realize this via their convolutional kernels and larger kernels can perform both. In Vision Transformers or other attention-based architecture, self-attention layers allow both (i) and (ii). Inspired by~\cite{tolstikhin2021mlp}, we design the CrossMLP architecture which firstly conducts cross-location (token-mixing) operations (ii) for one kind of information only, then conduct per-location (channel-mixing) operations (i) for both kinds of information. Note both (i) and (ii) are implemented with only MLPs in our CrossMLP and we summarize the architecture of our CrossMLP module in Figure~\ref{fig:crossblock}. 

Specifically, given two input features $F_{t-1}^{I}$ and $F_{t-1}^{S}$ with spatial shape $H \times W$ ( $H$ and $W$ indicate the height and width of a given feature). Both of them are firstly embedded to a sequence of $S$ non-overlapping feature map patches ($S = HW/P^{2}$, $P$ is the patch size) and each one is projected to a desired hidden dimension $C$ for generating the 2D real-valued input table $X_{I}{\in}{\mathbbm{R}^{S \times C}}$ and $X_{S}{\in}{\mathbbm{R}^{S \times C}}$.
Each CrossMLP consists of $7$ layers and each layer consist of two MLP blocks. The first one is the token-mixing MLP while another is the channel-mixing MLP. The former acts on columns of $X_{I}$ and $X_{S}$, which is similar to the original MLP-Mixer proposed by Tolstikhin \textit{et al.}~\cite{tolstikhin2021mlp} as follows:
\begin{equation}
\begin{aligned}
U_{*, i}^{I} & = X_{*,i}^{I} + W_{2}^{I}\sigma(W_1^{I}LayerNorm(X^{I})_{*, i}),\ for\ i = 1, \cdots, C\\
U_{*, i}^{S} & = X_{*,i}^{S} + W_{2}^{S}\sigma(W_1^{S}LayerNorm(X^{S})_{*. i}),\ for\ i = 1, \cdots,  C
\label{eq:1}
\end{aligned}
\end{equation}
The latter is shared across all rows but accompanied with another code by addition in a mutually cross interactive manner shown in Figure~\ref{fig:crossblock}. This operation can be described as follows:
\begin{equation}
\begin{aligned}
Y_{j, *}^{I} &= U_{j,*}^{I} + W_{4}^{I}\sigma(W_3^{I}LayerNorm(U^{I} + U^{S})_{j, *}),\ for\ j = 1, \cdots, S\\
Y_{j, *}^{S} &= U_{j,*}^{S} + W_{4}^{S}\sigma(W_3^{S}LayerNorm(U^{S} + U^{I})_{j. *}),\ for\ j = 1, \cdots,  S
\label{eq:2}
\end{aligned}
\end{equation}
Here $\sigma$ means element-wise nonlinearity (GELU \cite{hendrycks2016gaussian}), $W$ indicates the linear learnable weight of MLP. Based on this fashion, each kind of the information will be processed by one MLP layer along with the columns first, then each processed information accompanied with another information will be further strengthened mutually by the second MLP layer along with the rows. 

\noindent \textbf{Image Pathway Update.}
After we get the latent cues $Y^{I}$ and $Y^{S}$ about where to sample source patches and where to put patches. We design the image pathway update to let the cues guide the coarse image generation. Firstly, both $Y^{I}$ and $Y^{S}$ will be recovered from sequence shape $S \times C$ to the image shape $H \times W$, then a concatenation operation is adopted to combine them together. These steps can be depicted as follows:
\begin{equation}
\begin{aligned}
Y = cat(reshape(Y^{I}), reshape(Y^{S})).
\label{eq:3}
\end{aligned}
\end{equation}
Then $Y$ is transferred to an attention map denoted as $M_{Att}$, which incorporates both the image code at source view and semantic map code at target view. The value of $M_{Att}$ is in the range of $(0, 1)$ and indicates the importance of every element in the image code. Mathematically:
\begin{equation}
\begin{aligned}
M_{Att} = Sigmoid(Conv(Y)),
\label{eq:4}
\end{aligned}
\end{equation}
where $Conv$ is a convolutional layer with kernel size $1 \times 1$ to fully exchange information among channels. $Sigmoid$ here is to map the value after the convolutional layer to the range of $(0, 1)$. Finally, we update the image code as follows:
\begin{equation}
\begin{aligned}
F_{t}^{I} = F_{t-1}^{I} + M_{Att} \otimes Conv(F_{t-1}),
\label{eq:5}
\end{aligned}
\end{equation}
where $\otimes$ denotes element-wise multiplication, which ensures image code $F_{t}^{I}$ at one certain location are either preserved or suppressed. Moreover, the addition to $F_{t-1}^{I}$ constructs a residual connection~\cite{he2016deep} helps preserve the original image appearance information. It is critical especially when the number of the CrossMLP block goes higher.

\noindent \textbf{Semantic Map Pathway Update.}
To make the semantic map code can match the updated image code at the same feature level, we adopt channel-wise concatenation between $F_{t}^{I}$ and $Y$ as follows:
\begin{equation}
\begin{aligned}
F_{t}^{S} = cat(F_{t}^{I}, Y).
\label{eq:6}
\end{aligned}
\end{equation}
By doing so, both image code and semantic map code are activated mutually for generating more detailed images at the target view. 

\subsection{Refined Pixel-level Loss}
This noisy semantic label problem has been investigated in~\cite{kendall2018multi}, and it showed that weighing multiple loss functions by considering the uncertainty of each task can add regularization to the overall optimization objectives. This strategy was also adopted by SelectionGAN~\cite{tang2019multi}, which constructed 4 "tasks" with 4 fake-real pairs forming the uncertainty-guided pixel-level loss $\mathcal{L}_{p}$ to overcome this problem as follows:
\begin{equation}
	\begin{aligned}
		&\mathcal{L}_{p} \leftarrow  \frac{\mathcal{L}_{p}(Fake, Real)}{U}+ \log U = 
		\frac{\lvert Fake - Real \rvert}{U}+ \log U,
	\end{aligned}
	\label{eqn:6}
\end{equation}
where $\mathcal{L}_p$ denotes a pixel-level loss and $U$ denotes the relevant uncertainty map. $Fake$ and $Real$ indicate generated results and ground truth items.  We notice that in the SelectionGAN \cite{tang2019multi}, it need to compute the original pixel-level loss $\mathcal{L}_p$ 4 times for 4 $(Fake, Real)$ pairs, there are $(I_{g}^{'}, I_g)$, $(I_{g}^{''}, I_g)$, $(S_{g}^{'}, S_g)$, and $(S_{g}^{''}, S_g)$. We argue this approach is complicated and redundant. Moreover, 
it ignores the relation between generated image pair $(I_{g}^{'}, I_{g}^{''})$ and semantic map pair $(S_{g}^{'}, S_{g}^{''})$. 

Hence, we grouped these 4 fake-real pairs into 2 kinds based on whether the pair is image related or semantic map related. This makes the form of the loss become more compact than the one used in ~\cite{tang2019multi}. In addition, we add a new square item to constraint the difference between generated items for both stages. The refined pixel-level loss for both $\mathscr{L}_{g}$ and $\mathscr{L}_{s}$ share the same form as :
\begin{equation}
	\begin{aligned}
		&\mathcal{L}_{p} \leftarrow  
		\frac{\lvert Fake^{'} - Real \rvert + \lvert Fake^{''} - Real \rvert  }{U}+ \log U + (Fake^{'} - Fake^{''})^{2},
	\end{aligned}
	\label{eqn:7}
\end{equation}
The first item of our refined loss can reduce the distance between generated and the ground truth images for both stages at once, while the third square item further strengthen the consistency between the generated results from both stages.

\subsection{Model Training}
\noindent \textbf{Parameter-Sharing Discriminator.} We extend the vanilla discriminator original in~\cite{isola2017image} to another parameter-sharing structure. In the first stage, this structure takes the real image $I_a$ and the generated image $I_g^{'}$ or the ground-truth image $I_g$ as input. The discriminator $D$ learns to tell whether a pair of images from different domains is associated with each other or not. In the second stage, it accepts the real image $I_a$ and the generated image $I_g^{''}$ or the real image $I_g$ as input. This pairwise input encourages $D$ to discriminate the diversity of image structure and capture the local-aware information.

\noindent \textbf{Overall Optimization Objective.}
The total optimization loss is as:
\begin{equation}
\begin{aligned}
\min_{\{G_i, G_s, G_a\}} \max_{\{D\}} \mathcal{L} = & \sum_{i=1}^2 \lambda_i \mathcal{L}_{p}^i + \mathcal{L}_{cGAN} + \lambda_{tv}\mathcal{L}_{tv},
\end{aligned}
\label{eqn:all}
\end{equation} 
where $\mathcal{L}_{p}^i$ indicate our proposed refined pixel-level loss for both image and semantic map information, $\mathcal{L}_{cGAN}$ is the classic adversarial loss as follows:

\begin{equation}
\begin{aligned}
\mathcal{L}_{cGAN}\left(I_{a}, I_{g}^{(\cdot)}\right)=& \mathbb{E}_{I_{a}, I_{g}^{(\cdot)}}\left[\log D\left(I_{a}, I_{g}\right)\right]+ \mathbb{E}_{I_{a}, I_{g}^{(\cdot)}}\left[\log \left(1-D\left(I_{a}, I_{g}^{(\cdot)}\right)\right)\right],
\end{aligned}
\label{eqn:all}
\end{equation} 
where $I_g^{(\cdot)}$ means both the result $I_g^{'}$ from the first stage and $I_g^{'}$ from the second stage share the same adversarial loss. $\mathcal{L}_{tv}$ is the total variation regularization~\cite{johnson2016perceptual} on the $I_g^{''}$ as follows:
\begin{equation}
\begin{aligned}
\mathcal{L}_{tv}(I^{''}_{g})=\sum_{m, n}\left|I^{''}_{g[m+1, n]}-I^{''}_{g[m, n]}\right|+\left|I^{''}_{g[m, n+1]}-I^{''}_{g[m, n]}\right|,
\end{aligned}
\label{eqn:all}
\end{equation} 
where $m$ and $n$ indicate the coordinate of the pixels of $I_g^{''}$. $\lambda_i$ and $\lambda_{tv}$ are the trade-off parameters to control the relative importance of different objectives, and they are set to $0.5$ and $1$, respectively. The training is performed by solving the min-max optimization problem.

\noindent \textbf{Implementation Details.} Besides the proposed cascaded CrossMLP sub-network, we also utilize the selection module proposed in \cite{tang2019multi}. In addition, $G_s$ is a shallow U-Net \cite{isola2017image} architecture based generator and the filters in the first convolutional layer is set to 3. What's more, We adopt PatchGAN~\cite{isola2017image} for the discriminator $D$. About the training details,
we follow the optimization method in~\cite{goodfellow2014generative} to optimize the proposed Cascaded Cross MLP-Mixer GAN, i.e., one gradient descent step on discriminator and generators alternately.
We first train cascaded CrossMLP sub-network, selection module, and $G_s$ with $D$ fixed, and then train $D$ with others fixed.
Our method is trained and optimized in an end-to-end fashion. We employ Adam \cite{kingma2014adam} with momentum terms  $\beta_1{=}0.5$ and $\beta_2{=}0.999$ as our solver. The initial learning rate for Adam is 0.0002. The network initialization strategy is Xavier \cite{glorot2010understanding}, weights are initialized from a Gaussian distribution with standard deviation 0.2 and mean~0.

\section{Experiments}
\label{sec:exp}
\noindent \textbf{Datasets.}
We perform all experiments on two
datasets: (i) For the Dayton dataset~\cite{vo2016localizing}, we follow the same setting of~\cite{regmi2018cross} and select 76,048 images and create a train/test split of 55,000/21,048 pairs. And we reset the resolution to $256 \times 256$; (ii) The CVUSA dataset~\cite{workman2015wide} consists of 35,532/8,884 train/test image pairs. Following~\cite{zhai2017predicting,regmi2018cross}, the aerial images are center-cropped to $224 \times 224$ and resized to $256 \times 256$. For the ground-level images and corresponding segmentation maps, we also resize them to $256 \times 256$.

\begin{table*}[!t] \small
	\centering
	\caption{Quantitative comparisons on the Dayton dataset.
	}
	\resizebox{1\linewidth}{!}{%
		\begin{tabular}{cccccccccccccc} \toprule
			\multirow{2}{*}{Method}  & \multicolumn{4}{c}{Accuracy (\%) $\uparrow$}& \multicolumn{3}{c}{Inception Score $\uparrow$} & \multirow{2}{*}{SSIM $\uparrow$} & \multirow{2}{*}{PSNR $\uparrow$} & \multirow{2}{*}{SD $\uparrow$} & \multirow{2}{*}{KL $\downarrow$} \\ \cmidrule(lr){2-5} \cmidrule(lr){6-8} 
			& \multicolumn{2}{c}{Top-1} & \multicolumn{2}{c}{Top-5} & all & Top-1 & Top-5  \\ \hline
			Pix2pix \cite{isola2017image} &6.80$^\ast$ &9.15$^\ast$ &23.55$^\ast$&27.00$^\ast$ & 2.8515$^\ast$&1.9342$^\ast$&2.9083$^\ast$ & 0.4180$^\ast$ &17.6291$^\ast$ &19.2821$^\ast$ & 38.26 $\pm$ 1.88$^\ast$  \\ 
			X-Fork \cite{regmi2018cross}       &30.00$^\ast$&48.68$^\ast$&61.57$^\ast$&78.84$^\ast$& 3.0720$^\ast$&2.2402$^\ast$&3.0932$^\ast$ &0.4963$^\ast$&19.8928$^\ast$&19.4533$^\ast$  &6.00 $\pm$ 1.28$^\ast$ \\
			X-Seq \cite{regmi2018cross}       &30.16$^\ast$&49.85$^\ast$&62.59$^\ast$&80.70$^\ast$& 2.7384$^\ast$&2.1304$^\ast$&2.7674$^\ast$ &0.5031$^\ast$ &20.2803$^\ast$ &19.5258$^\ast$ & 5.93 $\pm$ 1.32$^\ast$ \\
			SelectionGAN~\cite{tang2019multi} & 42.11$^\dagger$ & 68.12$^\dagger$ & 77.74$^\dagger$ & 92.89$^\dagger$ & 3.0613$^\dagger$ & 2.2707$^\dagger$ & 3.1336$^\dagger$ & \textbf{0.5938}$^\dagger$ & \textbf{23.8874}$^\dagger$ & \textbf{20.0174}$^\dagger$ & 2.74 $\pm$ 0.86$^\dagger$ \\
			CrossMLP (Ours) & \textbf{47.65} & \textbf{78.59} & \textbf{80.04} & \textbf{94.64} & \textbf{3.3466} & \textbf{2.2941} & \textbf{3.3783} & 0.5599 & 23.6232 & 19.6688 & \textbf{2.33 $\pm$ 0.80} \\ \hline
			Real Data & - & - & - & - & 3.7196$^\dagger$ &  2.3626$^\dagger$ & 3.8998$^\dagger$ & - & - & - & - \\
			\bottomrule		
	\end{tabular}}
	\label{tab:Dayton}
	\vspace{-0.4cm}
\end{table*}

\begin{table*}[!t] \small
	\centering
	\caption{Quantitative comparisons on the CVUSA dataset.
	}
	\resizebox{1\linewidth}{!}{%
		\begin{tabular}{cccccccccccccc} \toprule
			\multirow{2}{*}{Method}  & \multicolumn{4}{c}{Accuracy (\%) $\uparrow$}& \multicolumn{3}{c}{Inception Score $\uparrow$} & \multirow{2}{*}{SSIM $\uparrow$} & \multirow{2}{*}{PSNR $\uparrow$} & \multirow{2}{*}{SD $\uparrow$} & \multirow{2}{*}{KL $\downarrow$} \\ \cmidrule(lr){2-5} \cmidrule(lr){6-8} 
			& \multicolumn{2}{c}{Top-1} & \multicolumn{2}{c}{Top-5} & all & Top-1 & Top-5  \\ \hline
			Pix2pix \cite{isola2017image} &7.33$^\ast$ &9.25$^\ast$ &25.81$^\ast$&32.67$^\ast$ & 3.2771$^\ast$&2.2219$^\ast$&3.4312$^\ast$ & 0.3923$^\ast$ &17.6578$^\ast$ &18.5239$^\ast$ & 59.81 $\pm$ 2.12$^\ast$  \\ 
			X-Fork \cite{regmi2018cross}           &20.58$^\ast$&31.24$^\ast$&50.51$^\ast$&63.66$^\ast$ &3.4432$^\ast$&2.5447$^\ast$&3.5567$^\ast$ & 0.4356$^\ast$ &19.0509$^\ast$ &18.6706$^\ast$ & 11.71 $\pm$ 1.55$^\ast$\\
			X-Seq \cite{regmi2018cross}             &15.98$^\ast$&24.14$^\ast$&42.91$^\ast$&54.41$^\ast$ &3.8151$^\ast$&2.6738$^\ast$&4.0077$^\ast$ & 0.4231$^\ast$ &18.8067$^\ast$ &18.4378$^\ast$ &15.52 $\pm$ 1.73$^\ast$  \\ 	SelectionGAN~\cite{tang2019multi} & 41.52$^\dagger$ & 65.51$^\dagger$ & 74.32$^\dagger$ & 89.66$^\dagger$ & 3.8074$^\dagger$ & 2.7181$^\dagger$ & 3.9197$^\dagger$ & \textbf{0.5323}$^\dagger$ & 23.1466$^\dagger$ & \textbf{19.6100}$^\dagger$ & 2.96 $\pm$ 0.97$^\dagger$\\
			CrossMLP (Ours) & \textbf{44.96} & \textbf{69.96} & \textbf{76.98} & \textbf{91.91} & \textbf{3.8392} & \textbf{2.8035} & \textbf{3.9757} & 0.5251 & \textbf{23.1532} & 19.5799 & \textbf{2.69 $\pm$ 0.94} \\ \hline
			Real Data & - & - & - & - & 4.8741$^\dagger$ &  3.2959$^\dagger$ & 4.9943$^\dagger$ & - & - & - & - \\
			\bottomrule		
	\end{tabular}}
	\label{tab:CVUSA}
	\vspace{-0.4cm}
\end{table*}

\noindent \textbf{Parameter Settings.}
For a fair comparison, we adopt the same training setup as in~\cite{isola2017image,regmi2018cross,tang2019multi} for both datasets. All images are scaled to $256{\times}256$, and we enabled image flipping and random crops for data augmentation. Similar to~\cite{regmi2018cross}, our method on the Dayton dataset is trained for 35 epochs. For the CVUSA dataset, we follow the same setup as in~\cite{zhai2017predicting,regmi2018cross}, and train our network for 30 epochs. Our method is implemented in PyTorch~\cite{paszke2019pytorch}.

\noindent \textbf{Evaluation Metrics.}
Inception Score, top-k prediction accuracy, and KL score are employed for the quantitative analysis. These metrics evaluate the generated images from a high-level feature space. We also employ  Structural-Similarity (SSIM), Peak Signal-to-Noise Ratio (PSNR), and Sharpness Difference (SD) to measure the pixel-level similarity. 

\subsection{State-of-the-Art Comparisons}
We compare our Cascaded Cross-Mixer GAN with four state-of-the-art methods Pix2pix~\cite{isola2017image}, X-Fork~\cite{regmi2018cross}, X-Seq~\cite{regmi2018cross}, and SelectionGAN~\cite{tang2019multi} both quantitatively and qualitative.

\begin{figure}[!t] \small
	\centering
	\includegraphics[width=0.8\linewidth]{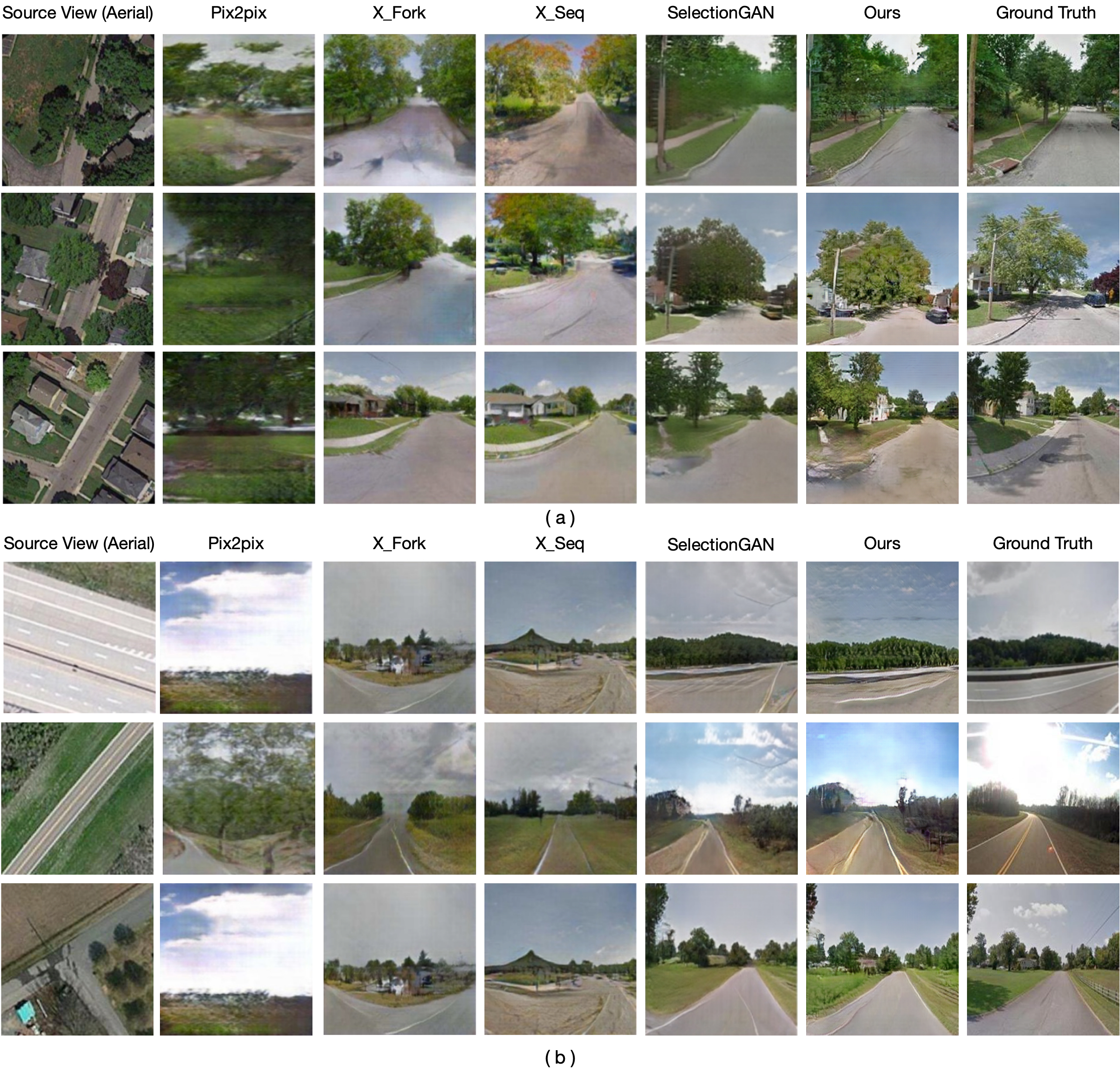}
	\caption{Qualitative results of different methods on (a) Dayton and (b) CVUSA datasets.
	}
	\label{fig:sota_results}
		\vspace{-0.4cm}
\end{figure}

\noindent \textbf{Quantitative Comparisons.}  
The results are shown in Tables~\ref{tab:Dayton} and~\ref{tab:CVUSA}. Note that ($\ast$, $\dagger$)  means these results are reported in~\cite{regmi2018cross} and \cite{tang2019multi}.
We can see the significant improvement of our method in both tables compared to other state-of-the-art methods on most of the metrics. Note that for metrics like SSIM, PSNR, and SD, our method also achieve the second-best results with only a bit lower on both datasets. However, the qualitative results presented in Figure~\ref{fig:sota_results} show our method can generate more realistic and sharper results than others. 

\noindent \textbf{Qualitative Comparisons.}
The qualitative results on both Dayton and CVUSA datasets are shown in Figure~\ref{fig:sota_results}. It's clear that the proposed Cascaded Cross-Mixer GAN can generate clearer and sharper scenes with more details but fewer artifacts on trees (see the first row in Figure~\ref{fig:sota_results}), clouds, roads(see the first and the fifth rows in Figure~\ref{fig:sota_results}), shadows (see the third row Figure~\ref{fig:sota_results}), sky (see the second and sixth rows in Figure~\ref{fig:sota_results}), than other methods. 

\subsection{Ablation Analysis}
To reduce the training time, we randomly select 1/3 samples from the whole Dayton dataset to form the ablation study dataset Dayton-Ablation with around 18,334/7,017 samples. We design five baselines (B1, B2, B3, B4, B5) as shown in Table~\ref{tab:ablation} and Table~\ref{tab:dayton_ablation_loss}. The first four baselines are used to figure out how the number of CrossMLP blocks affects the experimental results without the proposed refined loss. They use 3, 5, 7, and 9 blocks, respectively. The last one B5 is built on B4, but with our refined pixel-level loss for optimization during training. All the baseline models are trained and tested by the same configuration settings.

From the numerical results presented in Table~\ref{tab:ablation}. We conclude that with the increase of the number of cascaded CrossMLP blocks, the performance is also improved. What's more, from the comparison between B4 and B5 in Table~\ref{tab:dayton_ablation_loss}, it's clear that the refined pixel-level loss can boosts the numerical results obviously. We also provide more intuitive visual results in our supplementary materials. To further validate the effectiveness of the proposed refined pixel-level loss, another ablation experiment is conducted based on SelectionGAN~\cite{tang2019multi}, and results presented in Table~\ref{tab:dayton_ablation_loss} show that SelectionGAN can achieve much better performance on all the evaluation metrics with our proposed refined loss. 

In addition, Qiu \textit{et al.}~\cite{qiu2021semantic} also proposed a MLP based bilateral augmentation module (BAM) which is depicted somewhat similar to our CrossMLP module. Actually, they are totally different. Compared to our CrossMLP module, (1) For each module of BAM, only one input is from previous module and another is always the original 3D coordinates; (2) All MLP operations in BAM are in the channel-wise manner; (3) The cross interaction is realised by concatenation operation for two feature at different semantic level. Most importantly, we replace the CrossMLP module with the BAM (with slight modification to make it suitable for the data of our task) to see if it can achieve better performance based on B5. The results in Table~\ref{tab:dayton_ablation_loss} shows that our CrossMLP can outperform BAM by a large margin in all evaluation metrics. Hence we conclude the proposed CrossMLP module is more suitable and effective for the drastic cross-view image translation task.

\begin{table*}[!t] \small
	\centering
	\caption{Ablation results on the Dayton-Ablation dataset.}
	\resizebox{0.8\linewidth}{!}{%
		\begin{tabular}{ccccccccccc} \toprule
			\multirow{2}{*}{Method}  & \multicolumn{4}{c}{Accuracy (\%) $\uparrow$}& \multicolumn{3}{c}{Inception Score $\uparrow$} &  \multirow{2}{*}{KL $\downarrow$} \\ \cmidrule(lr){2-5} \cmidrule(lr){6-8} 
			& \multicolumn{2}{c}{Top-1} & \multicolumn{2}{c}{Top-5} & all & Top-1 & Top-5  \\ \hline
			B1 (3 Blocks) &43.68 & 72.84 &74.58 &91.23 & 3.2513 &2.2051 &3.2501 & 2.950 $\pm$ 0.90 \\ 
			B2 (5 Blocks)          &\textbf{44.35}&73.40 &\textbf{77.21}&91.60 &3.1825 &2.1625 &3.1889 & 2.803 $\pm$ 0.93 \\
			B3 (7 Blocks)             &42.58 &70.67 & 74.85 & 90.77 &3.2988 &2.2190 &3.3317  & 2.878 $\pm$ 0.91
			\\	
			B4 (9 Blocks) & 43.41 & 70.73 & 76.84 & \textbf{91.77} & 3.3241 & \textbf{2.2570} & \textbf{3.3340} & \textbf{2.725 $\pm$ 0.87}\\
            \hline
            Real Data & - & -&-&-&3.8246&2.5668&3.9119&-\\
			\bottomrule		
	\end{tabular}}
	\label{tab:ablation}
	\vspace{-0.4cm}
\end{table*}

\begin{table*}[!t] \small
	\centering
	\caption{Quantitative comparisons on the Dayton-Ablation dataset to validate the effectiveness of our proposed refined pixel-level loss.}
	\resizebox{1\linewidth}{!}{%
		\begin{tabular}{cccccccccccccc} \toprule
			\multirow{2}{*}{Method}  & \multicolumn{4}{c}{Accuracy (\%) $\uparrow$}& \multicolumn{3}{c}{Inception Score $\uparrow$} & \multirow{2}{*}{SSIM $\uparrow$} & \multirow{2}{*}{PSNR $\uparrow$} & \multirow{2}{*}{SD $\uparrow$} & \multirow{2}{*}{KL $\downarrow$} \\ \cmidrule(lr){2-5} \cmidrule(lr){6-8} 
			& \multicolumn{2}{c}{Top-1} & \multicolumn{2}{c}{Top-5} & all & Top-1 & Top-5  \\ \hline
			B4 &43.41 &70.73 &\textbf{76.84}&91.77 & 3.3241&2.2570&3.3340 & 0.4814 &21.0154 & 19.1099 & 2.725 $\pm$ \textbf{0.87}\\  
			B5 (B4 + Our Loss)&\textbf{44.08}&\textbf{73.46}&76.44&\textbf{92.28} &\textbf{3.4140}&\textbf{2.3197}&\textbf{3.4444} & \textbf{0.5515} &\textbf{22.9443} &\textbf{19.5192} & \textbf{2.699} $\pm$ 0.90\\
		    B5 + BAM~\cite{qiu2021semantic}&42.51&72.96&72.35&89.86&3.2756&2.1811&3.2703&0.5451&22.7198&19.4813&3.130 $\pm$ 1.03\\
			\hline
			SelectionGAN~\cite{tang2019multi} & 41.60& 71.21 & 75.20 & 92.01 & 3.2252 & 2.2387 & 3.2459& 0.4952 & 21.1408 & 19.3979 & 2.852 $\pm$ 0.93\\
			\cite{tang2019multi} + Our Loss & \textbf{42.75} & \textbf{71.47} & \textbf{75.70} & \textbf{92.39} & \textbf{3.2664} & \textbf{2.3101} & \textbf{3.2909}& \textbf{0.5442} & \textbf{21.1408} & \textbf{19.6984} & \textbf{2.704} $\pm$ \textbf{0.82}\\
			\hline
			Real Data & - & - & - & - & 3.8246 &  2.5668& 3.9119 & - & - & - & - \\
			\bottomrule		
	\end{tabular}}
	\label{tab:dayton_ablation_loss}
	\vspace{-0.4cm}
\end{table*}

\section{Conclusion}
\label{sec:conclu}
We propose a novel two-stage Cascaded Cross MLP-Mixer GAN to progressively address the challenging cross-view image translation task, particularly when there exits little or no overlap between source view and target view. In particular, our method can generate more detailed coarse results step by step in the first stage via our proposed cascaded CrossMLP blocks. What's more, within each block, the latent transformation cues can be well established by the CrossMLP module. Then the cross-view translation operation will be guided by those latent cues with better visual quality. In addition, we propose a refined pixel-level loss which can provide a further constraint between the coarse and final result images with a more compact formulation fashion, the inaccurate semantic labels problem can be also eased for better optimization with more reasonable regularization within the overall optimization objective. Extensive experimental results on two public datasets demonstrate that our method obtains much better results than the state-of-the-art.

\vspace{0.5cm}
\noindent \textbf{Acknowledgments.} \noindent This work was supported by the EU H2020 AI4Media No. 951911 project and by the  PRIN project PREVUE.

\bibliography{egbib}
\clearpage







This supplementary document provides additional experimental results of the proposed Cascaded Cross MLP-Mixer GAN for cross-view image translation task. We present the qualitative results of the ablation studies conducted in our main paper. 
\vspace{-0.5cm}
\section{Qualitative Results of Ablation Study}
\subsection{Effect of the Number of the CrossMLP Blocks}

\begin{figure}[!h] \small
	\centering
	\includegraphics[width=0.8\linewidth]{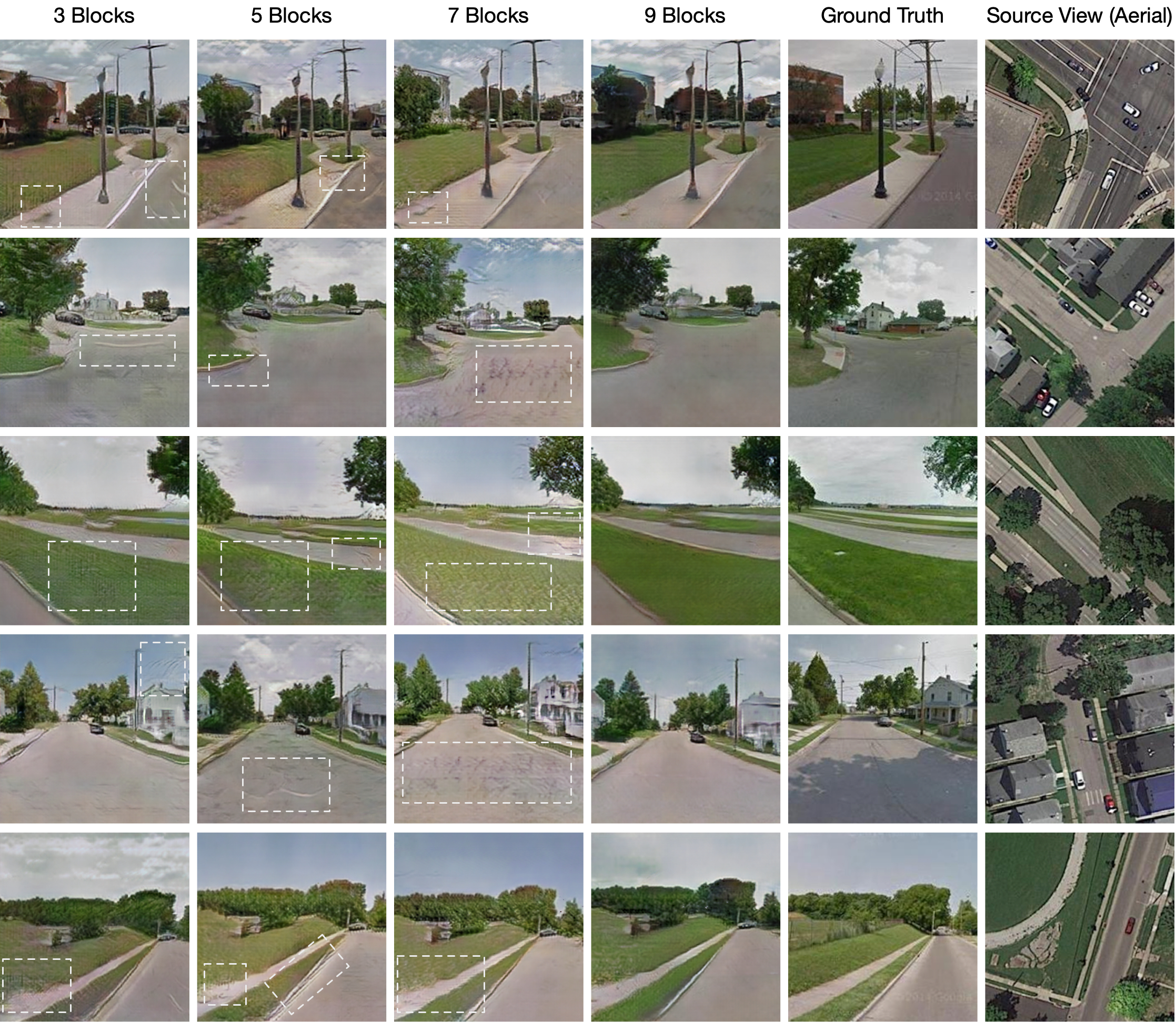}
	\caption{Visualization of the ablation study on the number of the CrossMLP blocks.
	}
	\label{fig:ablation_result}
	\vspace{-0.1cm}
\end{figure}

To figure out how the number of the proposed CrossMLP blocks affects the performance of the final results for our method. Besides the numerical results, we here present the visualization results in Figure~\ref{fig:ablation_result}. 

We can see that when the number of the CrossMLP blocks increases to 9, our proposed method achieves the best performance compared to others. We mark out those obvious artifacts for a better comparison. Particularly, the texture and color of grasses (in the third row), the smoothness of the road (in the forth row), and the color of the sky or clouds are more similar to the ground truth images. We conclude that with more CrossMLP blocks used in our experiments, both the geometry structure of a latent mapping pattern and the appearance information relevant to the details objects can be learned progressively. Hence, the final results can be more photorealistic than others. 

\begin{figure}[!t] \small
	\centering
	\includegraphics[width=0.8\linewidth]{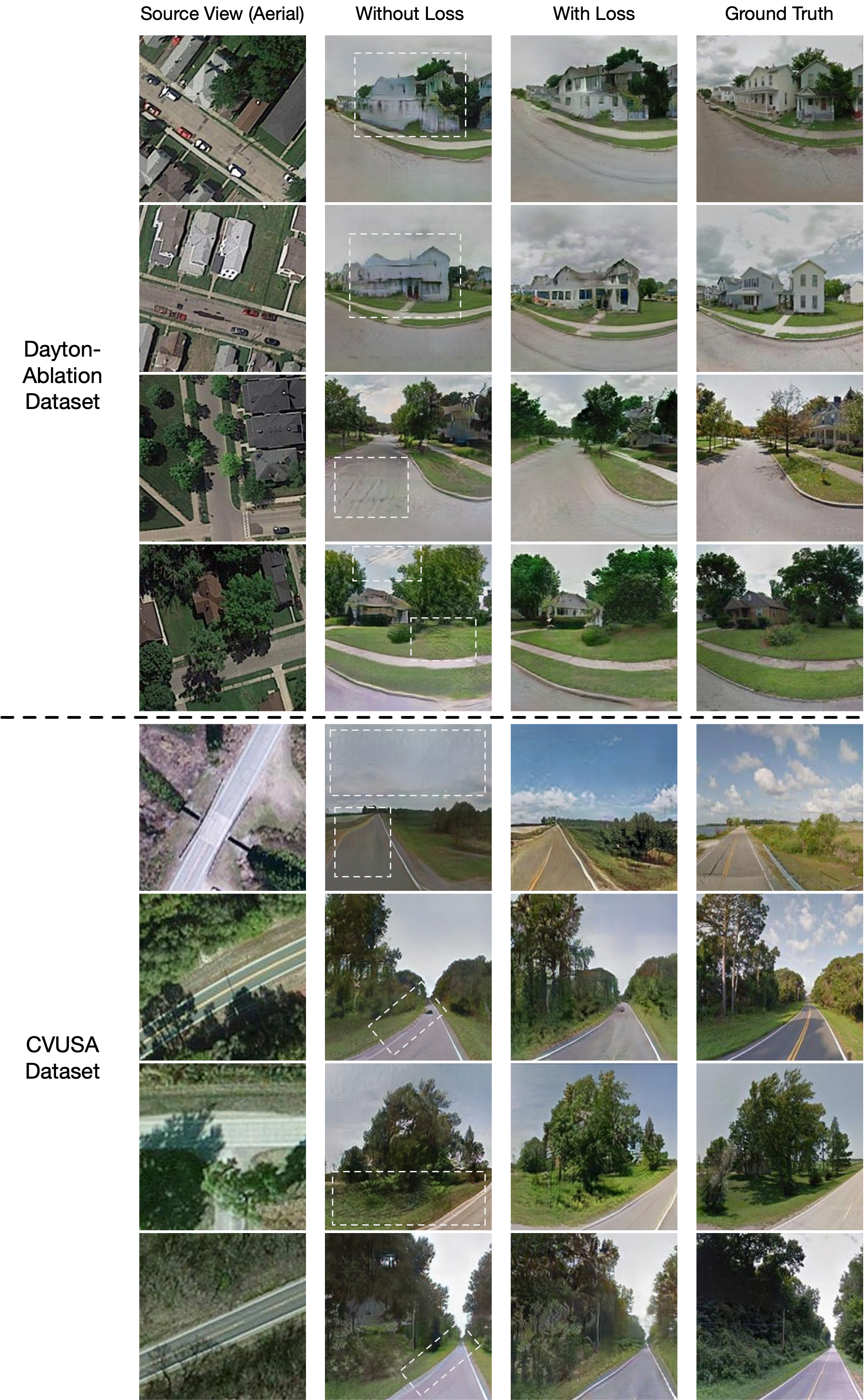}
	\caption{Visualization of the ablation study on the proposed refined pixel-level loss.
	}
	\label{fig:ablation_loss}
	\vspace{-0.4cm}
\end{figure}

\subsection{Effect of the Refined Pixel-Level Loss}
Besides the numerical comparison between B4 and B5 on Dayton-Ablation dataset in our main paper previously. We here provide more visualization results on both Dayton-Ablation and CVUSA dataset for giving further evidence to validate the superiority of our proposed method. In Figure~\ref{fig:ablation_loss}, we can see our method with the proposed refined pixel-level loss can generate more realistic house details (the first and the second row). In addition, the detailed things like the texture and style of road, trees, and grasses on Dayton-Ablation dataset becomes more photorealistic when we add the refined pixel-level loss. We can find the same trend on the CVUSA dataset, too. The color of grasses or sky, the lane line on the road, and the texture of trees or grasses are more photorealistic.

\end{document}